# Arabic Language Learning Assisted by Computer, based on Automatic Speech Recognition

**Naim TERBEH**[1]**, Mounir ZRIGUI**[2] [3]

[1]*Computer Science Department, Higher School of Science and Technology of Tunis*
[2]*LATICE Laboratory, Monastir, Tunisia*
[3]*Faculty of Sciences of Monastir, Tunisia*
terbehnaim1987@gmail.com  mounir.zrigui@fsm.rnu.tn

**Abstract** - This work consists of creating a system of the Computer Assisted Language Learning (CALL) based on a system of Automatic Speech Recognition (ASR) for the Arabic language using the tool CMU Sphinx3 [1], based on the approach of HMM. To this work, we have constructed a corpus of six hours of speech recordings with a number of nine speakers. we find in the robustness to noise a grounds for the choice of the HMM approach [2]. the results achieved are encouraging since our corpus is made by only nine speakers, but they are always reasons that open the door for other improvement works.

# 1. Introduction

The computer-assisted learning has benefited from the revolution of information technology caused by the emergence of multimedia computers. There are thousands of educational CD-ROM on market. The topics covered are vast: art, language learning, science, encyclopedias etc... Today, teaching by computer is on an extremely favorable terrain [24]. This article will attempt to show what is done and what can be done to integrate these tools of automatic speech processing in the software language learning by computer. We uses the technique of forced alignment to see the degree of correspondence between the model proposed by the teacher and the learner.

# 2. Work Context

This work takes place within the Research Laboratory Information and Communication Technology and Electrical Engineering (LATICE, Monastir unit). This article is part General of CALL, it is among the work of the computer-based Arabic language learning using automatic Arabic speech processing. We present in this work, at first, building the foundations of our speech recognition system. Next, we present our system of CALL in these parts of design, testing and discuss the results.

# 3. Arabic Language

Arabic is the language spoken originally by the Arabs. It is a Semitic language (as Akkadian and Hebrew). Within this set, it belongs to the Southern Semitic subgroup. Because of territorial expansion in the Middle Ages and the dissemination of the Koran, this language has spread throughout North Africa and Minor Asia. Arabic has 445 million speakers to be ranked fourth in the number of speakers, ranked eighth in number of pages that circulate on the Internet [4, 3].
By its morphological and syntactic proprieties, Arabic language is considered difficult to control in the field of automatic language processing [5,6]. this difficulty requires new methods that facilitate learning of Arabic, which is a CALL.

# 4. Existing Systems

For the most spoken languages in the world, several works on Assisted Computer Language Learning exist, we can mention:
- One of the first systems well documented is Plato (Programmed Logic for Automated Teaching Operations), developed at the University of Illinois and used for the teaching of Russian and other languages from 1970. As its name suggests, this authoring system is in the pedagogic movement of "teaching machines" of Skinner: is used to present information and generate structural exercises. In France, from 1969 and in the center "Computers for Education" of Paris VII, Francoise Demaizière develops software for English learning of English and devoted a thesis to this subject, including analyzing how enunciation of "dialogue" machine-learner. More recent work within the approach "Computer tutor" often stand in the behaviorist paradigm and hold a better hand account of developments in language teaching and have the other almost always an experimentation with learners, according to a paradigm psycholinguistics at most[7].
- An other is the system Kanda in Tokyo for English, German, French, Spanish and Chinese [8].
- Also note Mirto, LIDILEM developed in the laboratory of Grenoble. The project Mirto (Multi-Interactive Learning through Research on Texts and

Oral) is conducting a platform of language-teaching materials that use of linguistic material (for example: linguistics). MIRTO aims to put at the service teachers, resources and tools of NLP to enable them to design lesson plans exploiting the diversity and richness of the textual corpus. The scenarios will be developed are be used by learners from distance or face. The system should provide the possibility to calculate the "Profile" of each learner (mainly by exploiting its replies and its route), to provide an explanation of his mistakes and adapt scenarios to knowledge/skills in the course material for each learner[9]. Mirto is intended as a tool and environment serving language teachers, can allow them to express and implement proposals and solutions to problems relating to language teaching. We assume that these proposals and solutions can be conceived only through domain concepts. In this sense, any tool used must not only not to impose specific constraints, but allow easy manipulation of these concepts and facilitate their implementation [9].

- The project of the OPE [10], at the University of Paris VII, for the English language learning has a work of the CALL.

These achievements have undoubtedly a significant educational value, but they are limited by that they are applied only to the written language. we see that it is interesting to focus on the computer-assisted learning with automatic speech recognition, giving the student the ability to control his speaking. To our knowledge, little research has been done in this direction, we note the work of B. North-mann [11] at the University of Illinois, who proposed a project limited to the simulation of differences in pronunciation between teacher and student, and try to bring that possible the pronunciation of the learner to model proposed by the teacher.

Among the studies that are interested in Arabic, we can cite the work below:

- El-Kasasy [19] developed a system for learning the Koran holy. This system is based on segmentation of the signal in syllabic units. Each segment of the test is compared to the reference, then the system accepts or rejects the segment, it will not give detailed comments on the error.

- Omar proposed in [20] a system for identifying learner's pronunciation based on hidden Markov model (HMM). He brought together, in this work, different types of acceptable pronunciations of Arabic phonemes and compares them with the pronunciation's speaker to decide whether they are acceptable or not. The system involves two steps: First, the input pronunciation is segmented into phonemes. However, at this stage, substitution errors, insertion and deletion between phonemes of the search word is detected. Second, these units are processed by HMM.

We see that it is worth taking advantage of the automatic speech recognition for to the computer assisted foreign language learning.

## 5. Automatic Speech Recognition

The Automatic Speech Recognition (ASR) is a computer technology that allows a machine to extract the message oral contained in a speech signal. This technology uses computational methods in areas of signal processing and artificial intelligence. The What applications can imagine are many: help people with disabilities, control vocally machines, book flights, learn other languages, data entry, etc...

Most applications in RAP use technology based on Hidden Markov Models (HMM), which are capable of simultaneously modeling time and frequency characteristics of the voice signal. This technology offers high performance algorithms for learning and recognition, whereby HMM proved the best adapted to the problems of speech recognition. However they have limitations and difficulties, which lie mainly in the choice of a good initial model for learning and modeling of phoneme duration [12,13].

### 5.1. State of the art

The automatic speech processing opens up new perspectives in view of the considerable difference between the manual and classical command and voice. The use of the natural language in human-machine puts technology within reach of all and leads to its extension, reducing constraints on the use of keyboards, mice and command codes to master. By simplifying the protocol human-computer dialogue, the Automatic Speech Treatment is therefore also an increase in productivity since it is the machine that adapts to humans to communicate, and not vice versa. Moreover, it makes possible the use Simultaneous eyes or hands to another tasks. It allows to humanize computer systems of information management, focusing their design on users. Basically, the voice recognition software used primarily to enter text mass while spending keyboard (which offers a rate of 50 words per minute, against more than 150 per minute for speech) the keyboard remains still necessary corrections to text and use the computer.

### 5.2. Stages of recognition

The following diagram (Fig.1), present the process chain of automatic speech recognition.

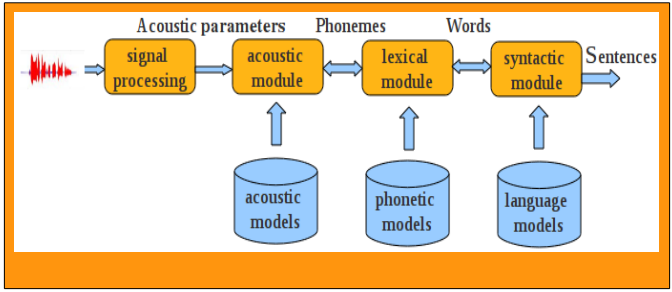

Figure 1. Process of automatic speech recognition

From a speech signal, the first treatment is to extract the parameters characteristics after pretreatment is done on frames of short duration (5 to 30 ms) used to improve the input signal, to reduce noise and eliminate the signal "not talk" (silence, music, etc ...). Then, to extract a feature vector (acoustic vector) for classification. The characteristic parameters generated are set in input of a module acoustic or an acoustic-phonetic decoding. The latter in turn can produce one or more phonetic hypotheses generally associated with a probability for each segment of the speech signal. This hypothesis generator is often modeled by local statistical models of elementary units of speech, for example a phoneme. To train the acoustic models, we learn models of acoustic units of a tagged corpus. The hypothesis generator interacts with a lexical module to force the acoustic-Phonetic decoding to recognizing only words represented in the lexical module. phonetic Models are represented by a dictionary of pronunciations (phonetic dictionary) or by the possibility of being represented by probabilistic automaton which are able to associate a probability to each possibility of word's pronunciation.

What sets the automatic recognition of continuous speech with large vocabulary compared to that of isolated words is that the hypothesis generator interacts with a module syntactic to force the recognition system to take into account the constraints syntactic or semantic. These constraints are formalized by language models. For recognize what is said, firstly, we look, using the models of acoustic units, unit which is supposed to have been produced, then build, from the succession of acoustic units and a statistical language model, the following words most probably.

The Bayesian equation applied to the topic of automatic speech recognition is formally summarized as follows: Let x be a sequence of acoustic vectors and unknown $w_i$ (i = 1 .. K) a K possible classes for this observation (phonemes, words, ...). Class recognized is:

$$w^* = \arg\max_i \frac{p(x/w_i).P(w_i)}{p(x)} = \arg\max_i p(x/w_i).P(w_i) \quad (1)$$

The recognized word w* will be the one that maximizes this quantity, among all candidate words $w_i$.

The probabilities $P(x|w_i)$ to observe the signal x given the sequence $w_i$ require a acoustic model to be estimated. The priori probability $P(w_i)$ of the sequence, regardless of signal, need a language model to be estimated. P(x) is the probability of the signal. It is the same for all possible sequences so its value is not taken into account.

## 5.3. Corpus

To make our system of automatic Arabic speech recognition, a corpus of multi-dialects voice is realized. This vocal database is in mono speaker with a 16khz sampling as shown in the following table (Table 1).

| Parameters | Values |
|---|---|
| Sampling | 16khz, 16 bit |
| wav format | Mono wav |
| Corpus | 5 hours of single words |
| Speakers | 8 (5 males + 3 females) |

Table 1. Parameters used for recording

The task of preparing corpus requires much attention to the selection of the words most representative from any domain and save them to the highest possible quality that allows a higher recognition rate. The most words used in the exchange of speeches are practically the same used for newspaper articles. Following this approach, we consulted the newspapers published in Arabic language to manually select the words more employees daily. We arrived to clear approximately 3500 words, most present in the Arabic spoken in various areas of daily life: politics, sports, economy, business, etc...

The recording of these words with three speakers, downloading the voice AASR corpus [14], and profit of the corpus realized by [15] allows us to reach almost 6 hours of recordings. The following table summarizes in detail the body's vocal resources of our work. in our work, we seek a little more robustness to the different Arabic dialects, which justifies the choice of AASR corpus with the dialect of golf country and Egypt.

| Speakers | Dialects | Duration |
|---|---|---|
| Speaker1 | Tunisian | 1 hour |
| Speaker2 | Tunisian | 1 hour |
| Speaker3 | Tunisian | 1 hour |
| Corpus[15] | Tunisian | 1h: 20 min |
| AASR's Corpus [14] | Golf/Egyptian | 1h: 30 min |
| Total | – | 5h: 50 min |

Table 2. Sources of speech corpus

The following table (Table 3) summarizes our corpus of work in statistical terms: learning and test.

| Corpus | Dialects | Duration | Male | Female |
|---|---|---|---|---|
| Learning | 3 | 4h:20min | 5 | 4 |
| Test | 3 | 1h:30min | 5 | 4 |
| Total | 3 | 5h:50min | 5 | 4 |

Table 3. Statistics of our corpus: Learning and Test

## 5.4. List of Arabic Phonemes

The phoneme is the unit that the decoder will recognize. We noted in Table 4 lists of the 44 phonemes used in our work. At the end of this list, we added the standard phoneme /SIL/ ([14], [21], [22]) which corresponds to periods of blank space between units of words.

| Nb. | Phoneme | Letter | Category |
|---|---|---|---|
| 1 | SIL | Silence | Silence |
| 2 | AE | فتحة مرققة | Short vowels |
| 3 | AA | فتحة مفخمة | |
| 4 | AH | فتحة مسبوقة بـ (*) | |
| 5 | UH | ضمة قصيرة | |
| 6 | UX | ضمة مفخمة | |
| 7 | IH | كسرة مرققة | |
| 8 | IX | كسرة مفخمة | |
| 9 | AE: | فتحة مرققة+ا | Long vowels |
| 10 | AA: | فتحة مفخمة+ا | |
| 11 | AH: | فتحة مسبوقة بـ(*)+ا | |
| 12 | UW | ضمة طويلة | |
| 13 | IY | كسرة طويلة | |
| 14 | IX: | كسرة مفخمة+ا | |

| Nb. | Phoneme | Letter | Category |
|---|---|---|---|
| 15 | DH2 | ظ | Emphatic (*) |
| 16 | TT | ط | |
| 17 | DD | ض | |
| 18 | SS | ص | |
| 19 | F | ف | Fricatives |
| 20 | TH | ث | |
| 21 | DH | ذ | |
| 22 | S | س | |
| 23 | Z | ز | |
| 24 | SH | ش | |
| 25 | KH | خ | |
| 26 | GH | غ | |
| 27 | AI | ع | |
| 28 | HH | ح | |
| 29 | H | ه | |

| Nb. | phoneme | letter | category |
|---|---|---|---|
| 30 | B | ب | Occlusive |
| 31 | T | ت | |
| 32 | D | د | |
| 33 | JH | ج | |
| 34 | K | ك | |
| 35 | Q | ق | |
| 36 | E | ء | |
| 37 | M | م | Nasal |
| 38 | N | ن | |
| 39 | R | ر | Liquid |
| 40 | L | ل | |
| 41 | W | و | Semivowels |
| 42 | Y | ي | |
| 43 | AW | وْ | |
| 44 | AY | يْ | |

Table 7. Shows details with the corresponding letter for each phoneme used

- Short vowels /AE/, /IH/ and /UH/ match diacritical marks in Arabic Fatha, Dhamma, Kasra.
- Long vowels are /AE :/, /IY/ and /UW/.
- The short vowel /AH/ is the diacritical mark of Arabic Fatha happens after an emphatic, that is long /AH:/.
- The short vowel / IX / corresponds to the diacritical mark of Arabic Kasra with Tafkhim, that is long /IX:/.
- The short vowel /AA/ corresponds to the diacritical mark of Arabic Fatha happens after a 'Tafkhim' (/R/ or /Q/), is the long /AA :/.
- The language vowel /UX/ corresponds to the Arabic diacritical mark of dhamma happens after a 'Tafkhim'.
- The phonemes /b/ and /D/ are similar to their English counterparts.
- The phonemes /M/ and /N/ are also similar to their English counterparts.
- /DD/ and /DH2/ correspond to the sounds of Arabic letters Thad (That as in English) and Dhad.
- /JH/ probe corresponds to the Arabic letter Jim.
- The phonemes /T/ and /K/ are fundamentally similar to their English counterparts.
- The sound of the letter occlusive Qaf is represented by the phoneme /Q/.
- The Hamza is represented by the phoneme /E/.
- Fricatives deaf Arab /F/, /S/, /TH/, /SH/ and /H/ are fundamentally similar to their English counterparts.
- Furthermore, the Arabic phonemes /SS/, /HH/ and /KH/ are the sounds: Sad, Hah, and Khah.
- The phonemes /AI/, /GH/, /Z/, and /DH/ correspond to the sounds: Ain, Ghain, Zain and Thel.
- The phonemes /Y/, /W/, /L/ and /R/ correspond to the sounds Yeh, Waw, Lam, and Reh.
- The phoneme /AY/ and /AW/ correspond to the sounds Y and W with a Sukun.

## 5.5. Pronunciation Dictionary

The dictionary is one of the textual components which is required for learning and recognition. It associates with each word one or more sequences of phonemes. So, one or several variations of acoustic pronunciations refer to the same written word. The quality of a automatic speech recognition system is highly dependent to the accuracy of pronunciation dictionary. The pronunciation dictionary must contain all words in the corpus to provide a important recognition rate, for that, we spent a significant period of our work, for build the dictionary. We had led finally to 25,841 words which is dictionary size sufficient to contain all the words in the corpus and ensure a significant recognition rate.

Creating a dictionary of pronunciations based necessarily on the list of Arabic phonemes noted in the previous session. The following figure (Fig. 2) shows an excerpt from the dictionary used in our work.

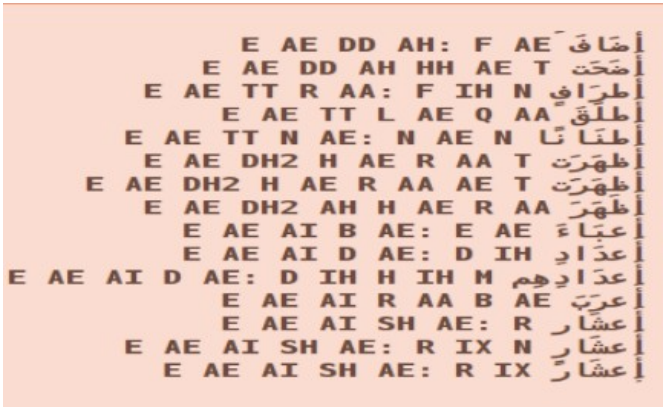

Figure 2. Extract the dictionary used

## 5.6. Acoustic-Phonetic Decoding

This process is responsible for generating, from a speech signal, the sequence of the phonemes more probably. This sequence of phonemes is based on parameter vectors extracted from the speech signal and assumptions provided by the acoustic model and the language model.

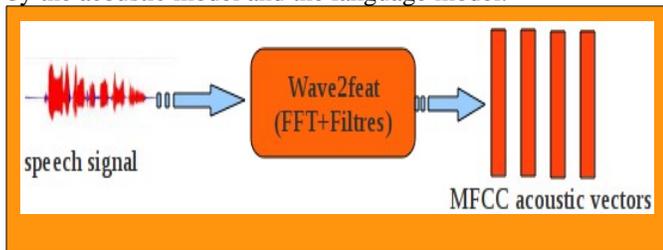

Figure 3. Speech signal to MFCC vector

The module consists of two sub models: the first is to generate the representative parameters of the signal (Fig. 3) because the decoder can't process signals in wav format so must first extract the MFCC parameters. The second, it's the decoding it self.
So, the necessary elements for the acoustic-phonetic decoding are:
– The MFCC parameters
– The Acoustic Model
– The language model
This is an illustration that best sums up the phenomenon of acoustic-phonetic decoding, in following figure (Fig. 4).

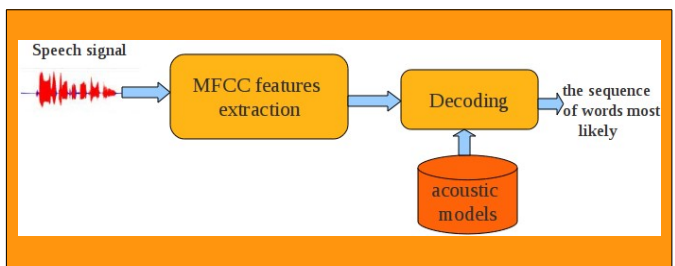

Figure 4. The acoustic-phonetic decoding

## 5.7. Principle of Forced Alignment

This treatment involves aligning speech signals with corresponding transcriptions, locate the phonetic description of the speech signal in the phonetic dictionary, identify the boundaries phonemic in the registration and assign a score to each phoneme depending on the distance separates from the norm which is the acoustic model.
We performed alignment with the Sphinx-align tool (in the box Sphinx3) who has possible to align the speech signal with the transcription. To do this, simply, we give the paths to the data signals that are necessary on MFCC format, the transcriptions, the phonetic dictionary and the acoustic model.

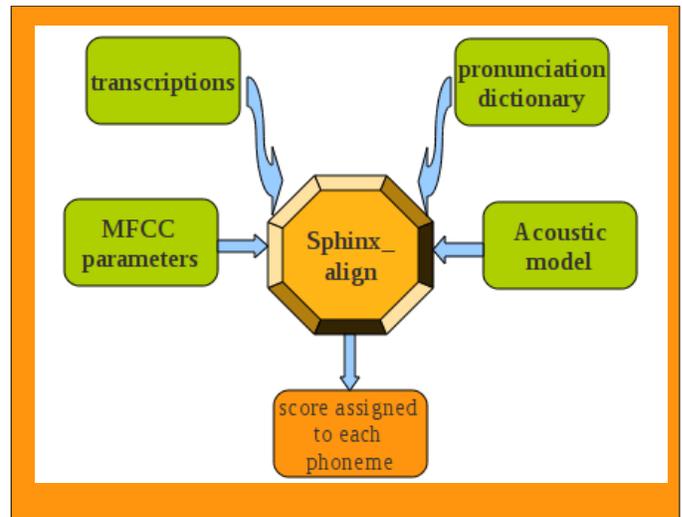

Figure 5. Forced alignment process

Once completed the forced alignment we get a segmented corpus into phonemes with results files that contain the start time and end time for each number of frames and the score which was affected, we find at the end of the file the sum of all scores for each signal.

## 5.8. Tests and Results

The purpose of evaluation of our recognition system we make the training and testing on speech corpus previously noted, the latter is characterized by multi-phrase: 5 male speakers and 4 female speakers. The results obtained are very satisfactory, indeed, we could achieve an important recognition rate. The results obtained are summarized as follows:

**Correct= 88.3%**
**WER= 11.7%**

With: Correct is the recognition rate,
WER is the error rate (Word Error Rate).
Respecting the following formulas:

$$\text{Percent Correct} = (N-D-S) / N \quad (2)$$
$$\text{Percentage WER} = 100-\text{Percent Correct} \quad (3)$$

With:
**N** the number of labels in the file transcription,
**D** the number of deletions,
**S** the number of substitutions.

The results obtained are very satisfactory given the size of our training corpus is relatively small with a reduced number of speakers. It is recommended that learning with nearly 500 different voices to reach a recognition rate of 100% [16]. we did not use a large corpus for learning, but our results are already encouraging.

## 5.9. Comparing and Discussing

To testing the effectiveness of our ASR system, we chose two other speech recognition for isolated words [17,18] and we made comparisons as the following table (Table 4) presents.

| Systems/criteria | Corpus's size (≈) | Approach | Recognition rate (%) |
|---|---|---|---|
| Our System | 5 hours | HMM | 88,3 |
| System1 | ¼ hours | Neural networks(NN) | 87 |
| System2 | ¼ hours | HMM | 84,45 |

Table 4. Comparison of Our System vs system1 and system2

The scene is full of much research work on the topic of speech recognition then our choice is justified by the type of recognition (recognition of isolated words for Arabic language). So, to ensure a logical of comparison, both systems are chosen for isolated words.

## 5.10. Conclusion

Seeing the recognition rate corresponding to each size of the corpus, although we note that it is not the primary factor that affects the results of a speech recognition system. So we must ensure a certain harmony between several factors: size's corpus, size of the pronunciation dictionary, recording quality (the least possible noise), the accuracy of the transcripts, good and correct pronunciation, proper approach, etc...

## 6. CALL Applications

The computer-assisted learning has attracted considerable attention in recent years [23]. Much research has been done to improve these systems particularly in the field of foreign language teaching. Along this section, we describe our system and the learning results of the Arabic spoken language assisted by computer. This work was developed to teach people (French) speak Arabic as foreign language, the correct pronunciation of phonemes. This application uses our automatic speech recognition system to detect errors in the pronunciation of the learner.

### 6.1. Design of our System

This system consists of a mosaic of sub-program managed by a main program that allows users to interact firstly with the teacher in the design phase of programmed instructions, and subsequently, with the student during the lesson itself. The operator-machine dialogue (teacher or student) is usually provided through a keyboard, screen and microphone.

#### 6.1.1. The role of the teacher:
the teacher should call the lesson and then:
- Select the words according to study problems of pronunciation adapted to grade level: Level 1 (A1), Level 2 (A2) or level 3 (B1).
- The teacher supervises the system and decide whether to grant the words that were chosen by him even for learners.

#### 6.1.2. The role of the learner:
learner "works" as follows:
- The student pronounces the words chosen by the teacher.

#### 6.1.3. The role of the system:
the system can produce results according to the pronunciation of the learner:
- If the pronunciation was incorrect, then the system returns the word after stressing the instead of faulty pronunciation.
- If the floor is too far from the model proposed by the teacher, especially if it is not provided that the fault that there will be only the error message.
- In the latter case one may ask:
  • To go directly to the hearing of the next word and continuing work.
  • Either repeat the word as many times if the teacher wants.

All these rules are designed to make our system a very simple application that provides a genuine dialogue with the student, even in the absence of the teacher.

### 6.2. Grouping of Phonemes

Phonemes present different challenges for each class, for learners. So, we find best to group them into classes as shown in Table 5.

| LSVG | WUS ح ع | SEOL ذ هـ ث خ ر | MFH | EL ق ص ض ط ظ | US |
|---|---|---|---|---|---|
| تكلَّم | حماية | خرج | سأل | طلب | في |
| أمِّي | حرب | هذا | سؤال | صفَّق | إسبانيا |
| تفَّاح | شارع | ماذا | وراء | قرص | لبنان |
| ستَّة | جامعة | ثمن | أمام | ظرف | سكن |

Table 5. grouping of Arabic phonemes

We explain here the abbreviation used in this table (Table 5)

LSVG : Long or short vowel germination
WUS: Words with unfamiliar sounds
SEOL: Sounds that exist in other languages
MFH : Middle and final "hamza"
EL: Emphatic Letters
US : Unproblematic Sounds

## 6.2. The process of CALL: Tests and Results

This paragraph addresses the test procedure of our system. This application was tested for sound information in its ability to provide statistics on a student and on a level. Systematic tests on a large Arabic corpus (from the order of 562 words) selected by a linguist and categorized according to the grouping in table 5.
The words are probably frequently used in daily life. They left as follows:
- Its not pose problems: 62 words.
- Letters emphatic: 100 words.
- Hamza in the middle and end: 100 words.
- Sounds existing in other languages: 100mots.
- Words with unfamiliar sounds: 100 words.
- Long vowel or short vowel and gemination: 100 words.

This system was tested by 13 French students from the Bourguiba Institute of Modern Languages, Tunisia, after learning of foreign languages, Arabic in our case. The following figures show the level statistics of each student for each class of Arabic phonemes.

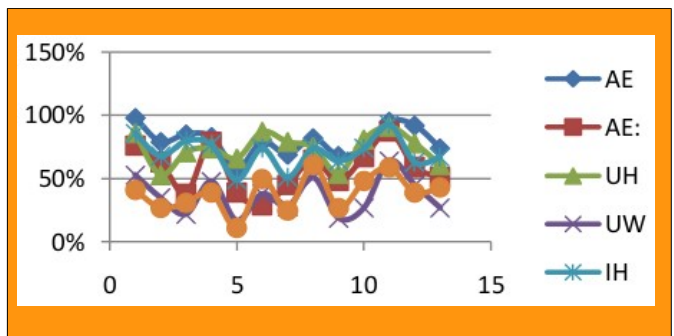

Figure 6. Results for long vowels, short vowels and gemination

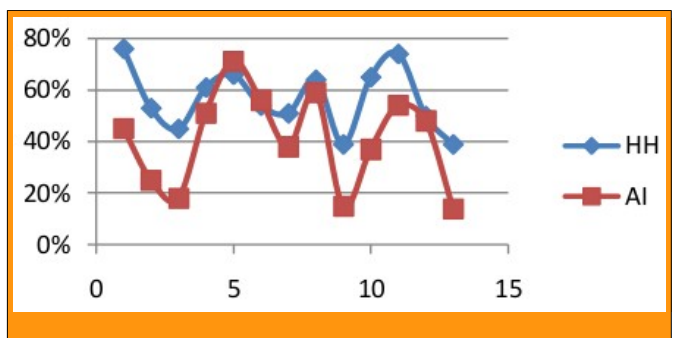

Figure 7. Results for the unknown sounds of Arabic

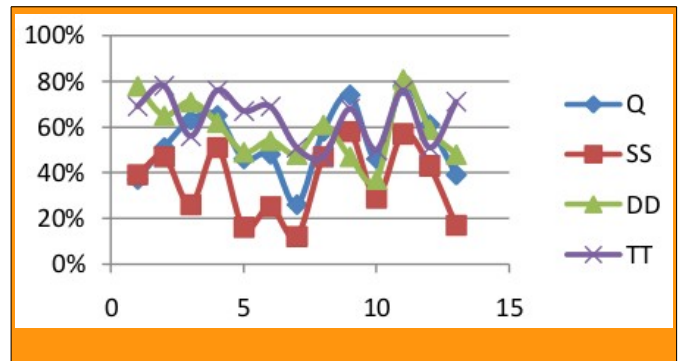

Figure 8. Results louse emphatic letters

## 6.3. Discuss

Note that the phonemes that exist in other languages do not have problems for learners at different levels, against the unknown phonemes cause pronunciation problems for most of the learners. The results for the emphatic letters (Figure 8) do not exceed the rate of 50% for at least 80% of learners so the teacher must more working on this side. Another problem of pronunciation for language vowels and gemination is been remarked upon seeing the results in Figure 6.

Generally, results and statistics of our system: spoken language learning: case of Arabic language, are very satisfactory. Previous statistics show levels of each student over his pronunciation difficulties of each class of phonemes. These statistics are very useful for the teacher to automatically detect errors in the pronunciation of each learner.

## 6.4. Conclusion

Analysis of forced alignment scores has us to give an idea of phonemes that cause problems for learners and the phonemes that are easier to master. We noticed that the sounds with characteristics similar to those of the mother tongue of learners are more difficult to master, in our case (Mother language: French, Foreign language: Arabic). Emphatic phonemes are the hardest to master. One of the difficult characteristics to control for learners is the duration of phonemes. We noticed this problem in the analyzes made of the short vowels and long vowels of Arabic and to gemination.

## 7. Future Works

During this work, we have not considered the level of each student which opens the focuses on a possible improvement based on the properties of each level (A1, A2, B1) and problems encountered by learners to another level. We find it interesting to make tests on all the phonemes of Arabic dialects [25], in more targeted at all levels even those that are considered advanced, and trying to incorporate more nationalities for learners. This is more than interesting as possible to improve the sound references and review corrective messages.

## 8. Conclusion

During this work, we got to build a corpus of six hours of speech recordings, we consult the operating mode of the recognition engine of Sphinx3 and make tests on this corpus. The results are satisfactory. In the second part, we presented our system of Computer Assisted foreign language Spoken learning: case of Arabic, it is based on the formalism of the Automatic Arabic Speech recognition. Our system has advantages over other work existing: it uses as a acoustic model of the Arabic speech based on hidden Markov model which gives results in the form of phonetic structures against by other works assume that the input signal is already phonetically labeled (as the case of El-Kasasy [19]). The help and correction messages show also a great advantage for our system.

## 9. References

[1]. http://cmusphinx.sourceforge.net.
[2]. J-P.HATON, C.Cerisara, D.Fohr, Y.Laprie, K.Smaiili, Reconnaissance Automatique de la Parole: du signal à son interprétation, Belgique, mai 2006.
[3]. SIL International, Ethnologue: Languages of the World, 15th Edition, ISBN 1-55671-159-X, 1272 pages, SIL International, Dallas, 2005.
[4]. http://fr.wikipedia.org/wiki/Liste_des_langues_par_nombre_total_de_locuteurs.
[5]. O. A LJLAYL , M. AND F RIEDER. On arabic search: Improving the retrieval effectiveness via a light stemming approach. In 11 the International Conference on Information and Knowledge Management (CIKM), pages 340–347, Virginia, USA, 2002.
[6]. L.S. L ARKEY, L. BALLESTEROS et M.E. CONNELL. Improving stemming for Arabic information retrieval: light stemming and co-occurrence analysis. In Proceedings of the 25th annual international ACM SIGIR conference on Research and development in information retrieval, pages 275 –282, Tampere, Finland, 2002.
[7]. Paru comme: Mangenot, F. (2005) «Seize ans de recherches en apprentissage des langues assisté par ordinateur». In Plurilinguisme et apprentissages, Mélanges Daniel Coste, p. 313-322. Lyon, ENS Editions.
[8]. Simon J. « L'éduction et l'informatisation de la société » Rapport au président de la république. 1981.
[9]. Georges Antoniadis, Claude Ponton , MIRTO : un système au service de l'enseignement des langues , LIDILEM – Université Stendhal de Grenoble, France , UNTELE 2004
[10]. Bestougeff H. Thèse de l'état université de Paris VII, 1970.
[11]. Nord-mann B. « A comparative study of some visual speech displays » Rapport de contract Université Illinois USA, 1981.
[12]. Aymen Trigui, Mohsen Maraoui, Mounir Zrigui: Acoustic Study of the Gemination Effect in Standard Arabic Speech. IPCV 2010: 192-196.
[13]. Aymen Trigui, Mohsen Maraoui, Mounir Zrigui: The Gemination Effect on Consonant and Vowel Duration in Standard Arabic Speech. SNPD 2010: 102-105.
[14]. http://www.ccse.kfupm.edu.sa/~elshafei/AASR.htm.
[15]. M-A.BenJannet, Construction d'un corpus vocal pour l'Arabe, PFE à l'unité de recherche LATICE, Monastir-Tunisie, juin 2010.
[16]. X.Huang, A. Acero, H. Hon, "Spoken language processing a guide to theory, algorithm and system design",Prentice Hall, 2001.
[17]. H.Satori, Système de Reconnaissance Automatique de l'arabe basé sur CMUSphinx.
[18]. N.Bakir, Reconnaissance Automatique des chiffres arabes en milieu réel par fusion audiovisuelle.
[19]. El-Kasasy M. « An Automatic Speech Verification System » Thèse, Cairo University, Faculty of Engineering Department of Electronics and Communications Egypt, 1992.
[20] Mourad Mars, Georges Antoniadis, Mounir Zrigui: Statistical Part Of Speech Tagger for Arabic Language. IC-AI 2010: 894-899.
[21]. M.Belgacem, Reconnaissance automatique de la parole et ALAO: Vers un système d'apprentissage de l'arabe oral, thèse de doctorat de l'université standhal-Grenoble3, décembre 2011.
[22]. Anis Zouaghi, Mounir Zrigui, Georges Antoniadis: Automatic Understanding of Spontaneous Arabic Speech - A Numerical Model. TAL 49(1): 141-166 (2008).
[23]. Mohsen Maraoui, Georges Antoniadis, Mounir Zrigui: CALL System for Arabic Based on Natural Language Processing Tools. IICAI 2009: 2249-2258.
[24]. Mohsen Maraoui, Georges Antoniadis, Mounir Zrigui: SALA: Call System for Arabic Based on NLP Tools. IC-AI 2009: 168-172.
[25]. Mohamed Belgacem, Mounir Zrigui: Automatic Identification System of Arabic Dialects. IPCV 2010: 740-749.